\documentclass[10pt,twocolumn,english]{IEEEtran}

\usepackage[T1]{fontenc}
\usepackage[latin9]{inputenc}
\usepackage[letterpaper]{geometry}
\usepackage{array}
\usepackage{booktabs}
\usepackage{multirow}
\usepackage{amsmath}
\usepackage{amssymb}
\usepackage{stackrel}
\usepackage{graphicx}
\PassOptionsToPackage{normalem}{ulem}
\usepackage{ulem}

\makeatletter

\providecommand{\tabularnewline}{\\}

\usepackage{algorithm}
\usepackage{algpseudocode}
\usepackage{flushend}

\usepackage[numbers,sort&compress]{natbib} 

\usepackage{breakurl}

\usepackage{colortbl}
\definecolor{grey}{cmyk}{0, 0, 0, 0.4}
\definecolor{lightblue}{cmyk}{0.25, 0.06, 0, 0.1}
\definecolor{palegreen}{cmyk}{0.39, 0, 0.39, 0.02}
\definecolor{lightgrey}{cmyk}{0, 0, 0, 0.17}

\makeatother

\usepackage{babel}
\begin{document}
\title{Trajectory Prediction for Autonomous Driving using Agent-Interaction
Graph Embedding}
\author{Jilan Samiuddin{*}, Benoit Boulet and Di Wu}

\maketitle
\noindent \footnote{This work was supported by Quebec\textquoteright s Fonds de recherche
Nature et technologies (FRQNT).\\The authors are with the Department
of Electrical and Computer Engineering at McGill University, Montreal,
QC, Canada\\{*}Corresponding author: jilan.samiuddin@mail.mcgill.ca}
\begin{abstract}
Trajectory prediction module in an autonomous driving system is crucial
for the decision-making and safety of the autonomous agent car and
its surroundings. This work presents a novel scheme called AiGem (Agent-Interaction
Graph Embedding) to predict traffic vehicle trajectories around the
autonomous car. AiGem tackles this problem in four steps. First, AiGem
formulates the historical traffic interaction with the autonomous
agent as a graph in two steps: (1) at each time step of the history
frames, agent-interactions are captured using spatial edges between
the agents (nodes of the graph), and then, (2) connects the spatial
graphs in chronological order using temporal edges. Then, AiGem applies
a depthwise graph encoder network on the spatial-temporal graph to
generate graph embedding, i.e., embedding of all the nodes in the
graph. Next, a sequential Gated Recurrent Unit decoder network uses
the embedding of the current timestamp to get the decoded states.
Finally, an output network comprising a Multilayer Perceptron is used
to predict the trajectories utilizing the decoded states as its inputs.
Results show that AiGem outperforms the state-of-the-art deep learning
algorithms for longer prediction horizons.
\end{abstract}


\section{Introduction}

Autonomous driving industry is advancing at a fast pace and demonstrating
its merits to alleviate several transportation challenges with regards
to safety, traffic-congestion, energy-saving, etc \cite{huang2022survey}.
Despite that, deployment of autonomous cars at a mass level is still
far from reality due to safety concerns. Human drivers continuously
predict the maneuvers of other vehicles on the road to plan a safe
and efficient future motion. Similarly, to ensure its own safety and
the safety of other agents on the road, the Autonomous Driving System
(ADS) must predict the motion of the surrounding agents in future
with high accuracy. High accuracy in prediction will also help improve
the capability of the autonomous car to infer future situations \cite{li2015real,bae2020cooperation}
and consequently enhance decision-making to enrich ride quality and
efficiency \cite{lin2021vehicle,phan2020covernet}. 

Trajectory prediction in real-time can pose significant challenges
given the dynamic and uncertain nature of roadways \cite{katariya2022deeptrack}.
Although traditional prediction models such as Kalman filter \cite{mourllion2005kalman},
car-following models \cite{sheng2020real}, and kinematic and dynamic
models \cite{polychronopoulos2007sensor,brannstrom2010model} are
hardware efficient, they become unreliable in long-term predictions
when the spatial-temporal interdependence is ignored \cite{lefevre2014survey}.
To mitigate this concern, Ju et al. \cite{ju2020interaction} combined
Kalman filter, kinematic models and neural networks and obtained better
performance than the traditional ones. Predicting multiple possible
trajectories and ranking them based on probability distribution of
the prediction model are also found in the literature \cite{deo2018convolutional,phan2020covernet}.
However, these approaches are inherently less pragmatic in real-time
scenarios \cite{katariya2022deeptrack}.

With significant developments in deep learning, particularly successful
implementations of long-short term memory (LSTM) in capturing temporal
dependencies, several works have been conducted such as \cite{zyner2018recurrent,xin2018intention}
for trajectory prediction. Despite obtaining high accuracy, these
techniques do not include the interactions between the vehicles. Deo
et al. \cite{deo2018convolutional} addresses this shortcoming by
modeling the spatial interactions between the vehicles with a social
tensor and then further extracting features using a convolutional
social pooling techniques. However, the social tensor only preserve
the spatial interaction of the last time stamp of the history, and
thus, the spatial-temporal relation is not captured.

Li et al. proposes GRIP++ \cite{li2019grip++} that capture the spatial-temporal
relation using fixed and dynamic homogeneous graphs. They offer two
different types of edges -- spatial and temporal -- to capture the
environmental dynamics. This is similar to what we propose, except
that we use a single heterogeneous graph to feed through the graph
convolutional model. Furthermore, an encoder GRU network is used in
\cite{li2019grip++}, whereas, in our proposed scheme, the graph convolutional
model acts as the encoder network. As a result, the model size of
GRIP++ is significantly larger than ours. Sheng et al. \cite{sheng2022graph}
also proposes a similar graph-based technique in which they stack
spatial graphs from each time step in the past to form a spatial-temporal
homogeneous graph. They use two different modules for extracting the
spatial dependency and then extract the temporal dependency. However,
since in our work, we include both the spatial and temporal aspects
in the construction of the graph itself, the graph convolutional model
alone is sufficient.

Lin et al. \cite{lin2021vehicle} uses spatial-temporal attention
mechanisms with LSTM for trajectory prediction with primary focus
on the explainability of their models. Katariya et al. \cite{katariya2022deeptrack}
focuses on model complexity and size for faster performance and lower
memory requirement. They use depthwise TCN-based (Temporal Convolutional
Networks) encoder instead of LSTMs reducing the overall model size.
The performances of both \cite{lin2021vehicle} and \cite{katariya2022deeptrack}
are comparable, even though not the best, to other existing state-of-the-art
methods. In our work, we integrate attention in the depthwise graph
convolutional network by graph attention networks (GAT) to capture
the importance of neighboring vehicles.

This article proposes a trajectory prediction method for autonomous
driving using agent-interaction graph embedding (AiGem). AiGem first
constructs a heterogeneous graph from the historical data that encapsulates
the interactions between the agents required for trajectory prediction
of these agents. A graph encoder network then processes the graph
to find embedding of the target vehicles at the current time stamp
which is fed through a sequential GRU encoder network for trajectory
prediction with the aid of an Multilayer Perceptron (MLP)-based output
network. The results show that AiGem has higher accuracy in longer
prediction horizons in contrast to the existing state-of-the-art methods
with comparable accuracy for lower predictions horizons. The size
of the model is much lighter than that of the compared methods except
for \cite{sheng2022graph}.

\section{Problem Formulation\label{sec:Problem-Formulation}}

The trajectory prediction module is responsible to estimate the future
positions of all the actors in the current frame based on their trajectory
histories given by their global coordinates $x$ and $y$ with respect
to the autonomous agent (also known as the ego) at the current time
step, their heading $\theta$, and the velocity $v$. Formally, the
trajectory histories over $\tau_{H}$ seconds, with a sampling time
$t_{s}$ (i.e., over $K_{H}=\frac{\tau_{H}}{t_{s}}+1$ time steps),
of all the observed actors can be shown as follows{\footnotesize{}
\begin{equation}
\mathcal{H}=\left[X_{1},X_{2},\ldots,X_{K_{H}}\right]
\end{equation}
}where, $X_{k}=\left[x_{k}^{1},y_{k}^{1},\theta_{k}^{1},v_{k}^{1},\ldots,x_{k}^{N_{k}},y_{k}^{N_{k}},\theta_{k}^{N_{k}},v_{k}^{N_{k}}\right]$
with $N_{k}$ being the number of actors observed at the $k^{\mathrm{th}}$
step. The task of the prediction module is to predict the future positions
over a prediction horizon $\tau_{F}$ (i.e., over $K_{F}=\frac{\tau_{F}}{t_{s}}$
time steps) of all the $N_{K_{H}}$ observed actors:{\footnotesize{}
\begin{equation}
\mathcal{F}=\left[Y_{K_{H}+1},Y_{K_{H}+2},\ldots,Y_{K_{H}+K_{F}}\right]
\end{equation}
}where, $Y_{k}=\left[x_{k}^{1},y_{k}^{1},x_{k}^{2},y_{k}^{2},\ldots,x_{k}^{N_{K_{H}}},y_{k}^{N_{K_{H}}}\right]$.
Note that position with respect to the ego at the current time step
$K_{H}$ infers that the position of the ego $\left(x_{K_{H}}^{e},y_{K_{H}}^{e}\right)$
at present is set to $\left(0,0\right)$ and the coordinate frame
is shifted as such. This, to some extent, prevents outliers for positional
features to occur during training that could lead to poor accuracy
\cite{khamis2005effects}. 

\section{Preliminaries\label{sec:Preliminaries}}

\subsection{Graph Neural Network\label{subsec:Graph-Neural-Network}}

Graphs are used to model the complex interactions between agents,
also called the nodes, and their involvement in a neighborhood. The
interactions between the nodes are presented using edges. Heterogeneous
graphs, unlike homogeneous graphs, are more complex and have nodes
and/or edges that can have various types or labels associated with
them, indicating their different roles or semantics. For example,
in our work, we use spatial and temporal edges requiring a heterogeneous
graph representation. Because of the differences in type, a single
edge feature tensor is unable to accommodate all edge features of
the graph \cite{Heterogeneous_PyG}. The computation of messages and
update functions is conditioned on node or edge type. We use PyTorch
libraries in our work to operate on heterogeneous graphs which are
detailed in \cite{Heterogeneous_PyG}.

Graph Neural Networks (GNNs) are designed for the precise task of
processing graph \cite{scarselli2008graph}. In the node representations
task, the GNN generates the matrix embedding of the nodes of size
$\mathbb{R}^{n\times m}$, where, $n$ is the number of nodes and
$m$ is the dimensionality of the features of the nodes. For node
$p$, the GNN aggregates messages from its neighbors and then applies
a neural network in several layers \cite{stanfordGNN}:{\footnotesize{}
\begin{equation}
h_{p}^{(i+1)}=\sigma\left(W_{i}\underset{q\in\mathcal{N}(p)}{\sum}\frac{h_{q}^{(i)}}{\left|\mathcal{N}(p)\right|}+B_{i}h_{p}^{(i)}\right),
\end{equation}
}$\forall i\in\left\{ 0,1,\ldots,L-1\right\} $, where, $h$ is the
embedding of a node, $\sigma$ is a non-linear activation function,
$W_{i}$ and $B_{i}$ are the weights and biases of the $i^{\mathrm{th}}$
layer, respectively, $\mathcal{N}(p)$ is the neighborhood of the
target node $p$, and $L$ is the total number of layers. Message
aggregation can be done, for example, by averageing the messages from
the neighbors of $p$.

Graph Attention Network (GAT) \cite{velivckovic2017graph} is a GNN
variant that integrates attention to focus on learning of the more
relevant aspects of the input. With continual aggregation of information
from the neighboring nodes $q$ of the target node $p$, the network
learns the importance of these neighboring nodes, also known as the
attention coefficient $e_{pq}$. The attention coefficient is computed
by applying a common linear transformation $\boldsymbol{W}$ to the
features ($\boldsymbol{h}$) of both $p$ and $q$, followed by a
shared attentional mechanism ($\mathrm{att}$) as follows:{\footnotesize{}
\begin{equation}
e_{pq}=\mathrm{att}\left(\boldsymbol{W}\boldsymbol{h}_{p},\boldsymbol{W}\boldsymbol{h}_{q}\right)
\end{equation}
}A single-layered feed-forward neural network is used for $\mathrm{att}$
in \cite{velivckovic2017graph}. For fair comparison between the neighboring
nodes, $e_{pq}$ is normalized using the softmax function:{\footnotesize{}
\begin{equation}
\alpha_{pq}=\frac{\mathrm{exp}\left(e_{pq}\right)}{\sum_{k\in\mathcal{N}(p)}\mathrm{exp}\left(e_{pk}\right)}\label{eq:attention_coeff}
\end{equation}
}\cite{velivckovic2017graph} has more detailed explanation on GAT
for interested readers.

\subsection{Gated Recurrent Unit (GRU)}

\begin{figure}[t]
\begin{centering}
\includegraphics[scale=0.05]{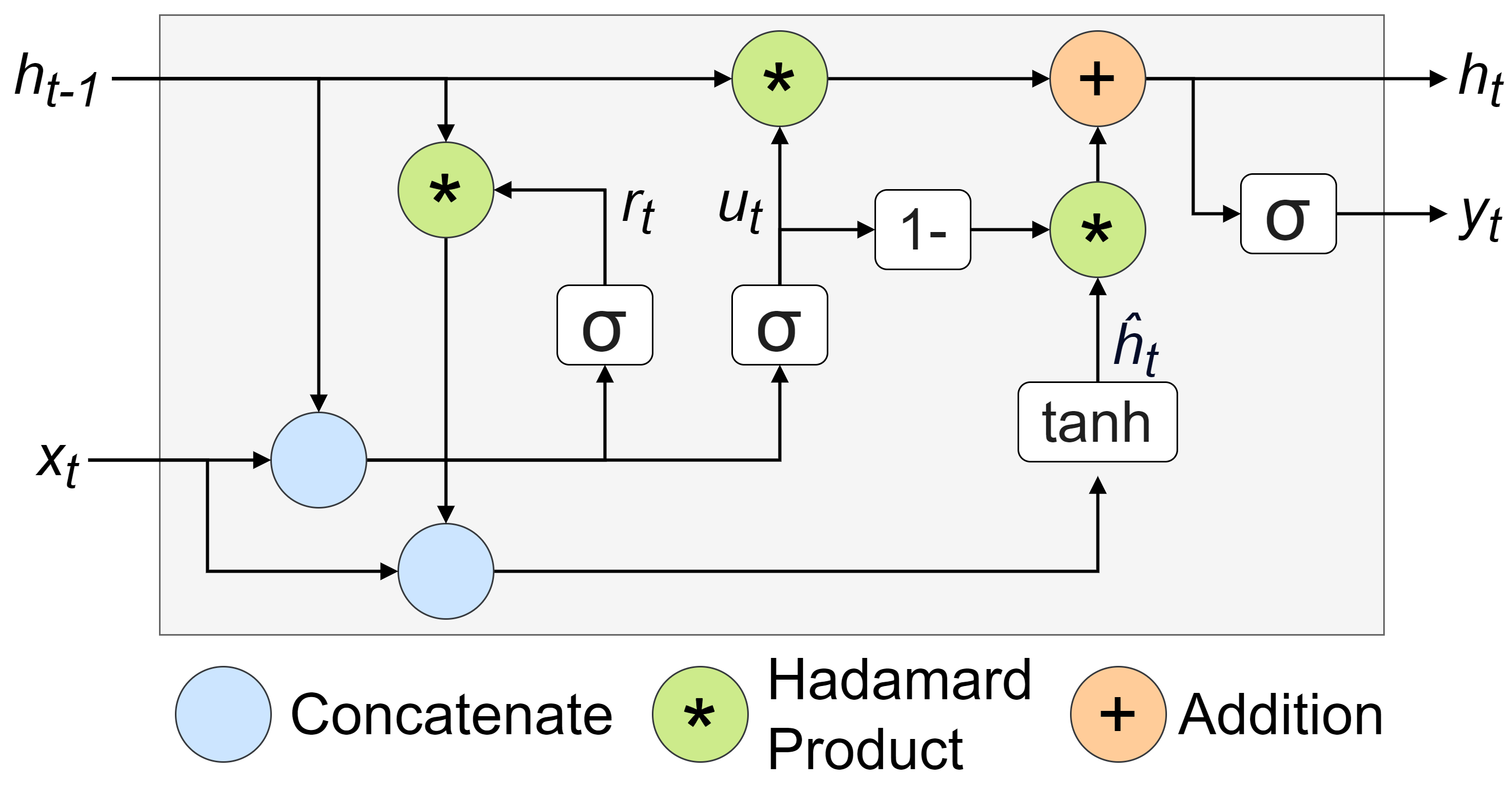}
\par\end{centering}
\begin{centering}
\vspace*{-3mm} 
\par\end{centering}
\caption{GRU architecture\label{fig:GRU-architecture}}
\end{figure}

GRU \cite{cho2014learning} aims to solve the vanishing gradient problem
typically found in a recurrent neural network (RNN). It does so by
adding an update gate $u_{t}$ and a reset gate $r_{t}$ which are
responsible for determining the flow of information to the output.
Figure \ref{fig:GRU-architecture} shows the architecture of GRU that
takes the current state $x_{t}$ and the previous hidden state $h_{t-1}$
as inputs. The current output state $y_{t}$ and the hidden state
$h_{t}$ are then obtained using the following:{\footnotesize{}
\begin{equation}
\begin{aligned}r_{t}= & \sigma\left(W_{rx}x_{t}+W_{rh}h_{t-1}\right)\\
u_{t}= & \sigma\left(W_{ux}x_{t}+W_{uh}h_{t-1}\right)\\
\hat{h}_{t}= & \mathrm{tanh}\left(W_{\hat{h}x}x_{t}+W_{\hat{h}h}\left(r_{t}*h_{t-1}\right)\right)\\
h_{t}= & u_{t}*h_{t-1}+\left(1-u_{t}\right)*\hat{h}_{t}\\
y_{t}= & \sigma\left(W_{y}h_{t}\right)
\end{aligned}
\label{eq:GRU}
\end{equation}
}where, $W_{rx}$, $W_{rh}$, $W_{ux}$, $W_{uh}$, $W_{\hat{h}x}$,
$W_{\hat{h}h}$, $W_{y}$ are weights, and the operator $*$ is the
Hadamard product. Updates are applied using equation (\ref{eq:GRU})
to obtain hidden states and outputs recurrently.

\section{Proposed Methodology}

\begin{figure*}[t]
\begin{centering}
\includegraphics[scale=0.04]{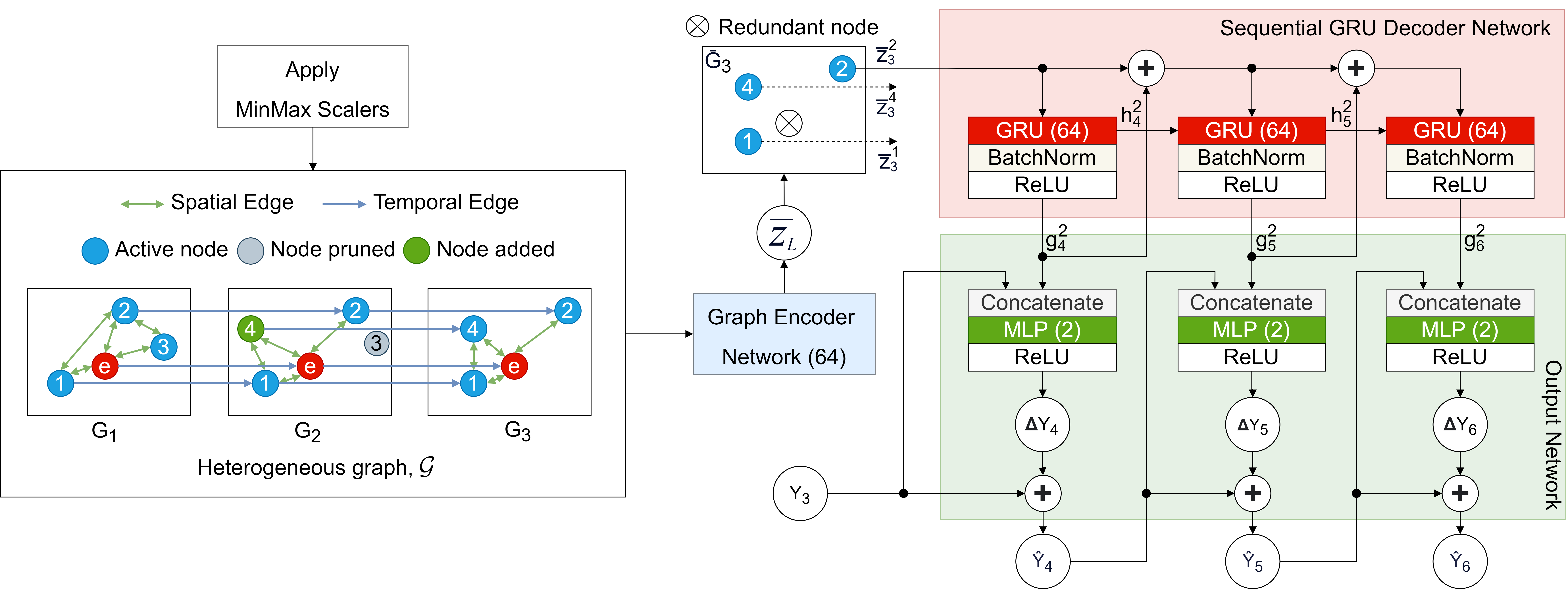}
\par\end{centering}
\begin{centering}
\vspace*{-3mm} 
\par\end{centering}
\caption{Proposed network architecture AiGem with $K_{H}=3$ (number of steps
in the history including the present), $K_{F}=3$ (number of steps
to predict). Modules (GRUs and MLPs) inside the networks with the
same color share the same weights. The number within parenthesis represents
the output dimension of the module. \label{fig:Network-Architecture}}
\end{figure*}

In this section, we propose a novel way to articulate the trajectory
prediction problem as a sequence of connected graphs based on the
historical data. Our proposed network architecture, as shown in Figure
\ref{fig:Network-Architecture}, illustrates the three network components
forming a coherent structure for solving the trajectory prediction.
The three components are (1) graph encoder network, (2) sequential
GRU decoder network, and (3) output network. 

\subsection{Graph Formulation\label{subsec:Graph-Formation}}

\begin{figure}[t]
\begin{centering}
\includegraphics[scale=0.045]{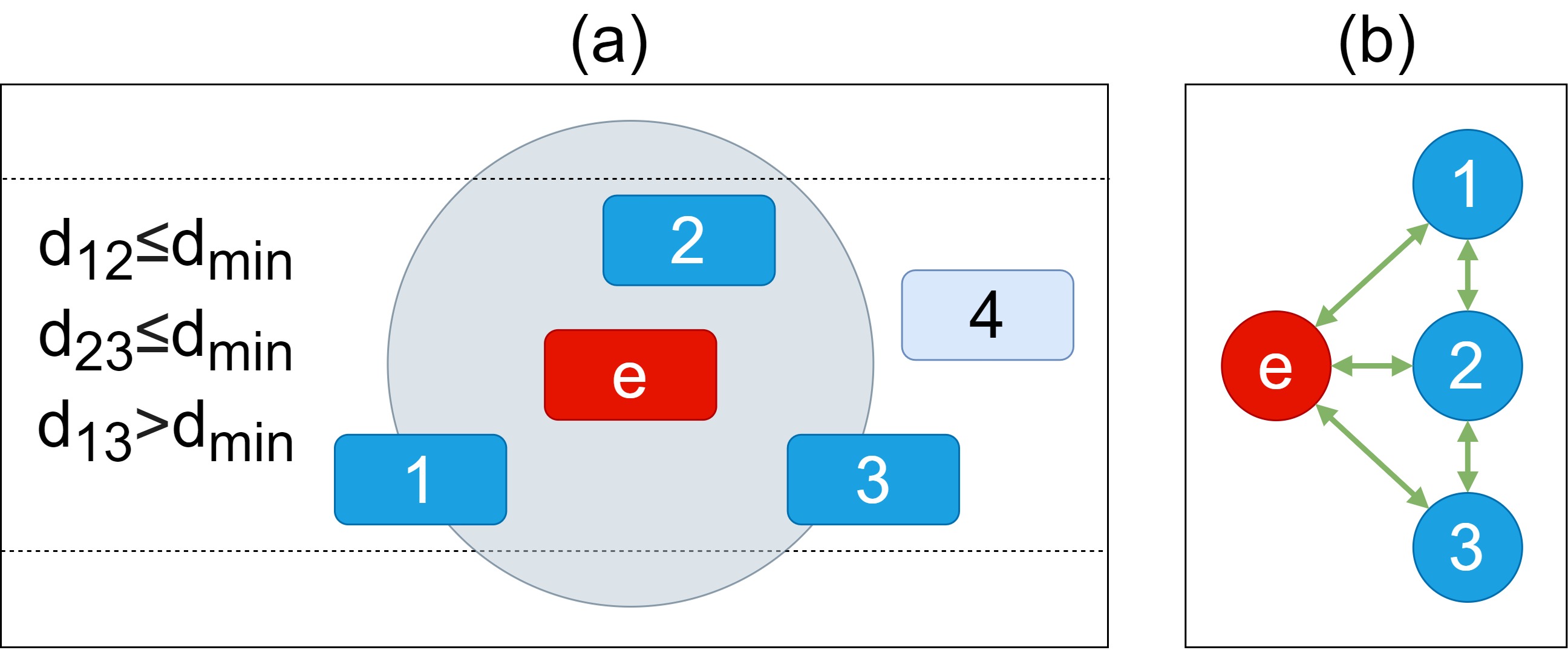}
\par\end{centering}
\begin{centering}
\vspace*{-3mm} 
\par\end{centering}
\caption{(a) An example scenario of an ego (in red) surrounded by actors (blue)
of which actors 1, 2, and 3 are in its sensing area (gray circle),
and, (b) Graph formulation with ego connected to the sensed actors
via bidirectional spatial edges. \label{fig:Graph-example}}
\end{figure}

First, the spatial graph $G_{k}$ for each time step in the past,
i.e., $k=1,2,\ldots,K_{H}$, is generated. Figure \ref{fig:Graph-example}(a)
shows an example scenario at the $k^{\mathrm{th}}$ time step where
the ego is surrounded by four actors of which three actors (1, 2 and
3) are in its sensing area, while the fourth actor is not in that
range. Note that the sensing area in this work is defined to be a
circle of radius around the ego. In the formulation of the graph as
shown in Figure \ref{fig:Graph-example}(b), alongside the ego (referred
to as node $e$), an actor $i$ is considered as a node of graph $G$
if the inequality on the right-hand side of the following equation
is met: {\footnotesize{}
\begin{equation}
\mathbb{I}_{\mathrm{node}}\left(i_{k}\right)=\begin{cases}
1, & \textrm{if }d_{ei}\leq50\\
0, & \textrm{otherwise}
\end{cases},
\end{equation}
}where, $\mathbb{I}$ is an indicator function and $d_{ei}$ is the
euclidean distance between the ego and the $i^{\mathrm{th}}$ actor.
Each node $e_{k},i_{k}\in\mathcal{N}_{k}$, where $\mathcal{N}_{k}$
is the set of all nodes present at the $k^{\mathrm{th}}$ time step,
has the following feature vector:{\footnotesize{}
\begin{equation}
z_{k}^{e/i}=\left[\begin{array}{cccc}
x_{k} & y_{k} & \theta_{k} & v_{k}\end{array}\right]\in\mathbb{R}^{4}
\end{equation}
}The ego $e$ always shares a bidirectional edge with a detected actor
$i$. 

{\footnotesize{}
\begin{equation}
\mathbb{I}_{\mathrm{ea}}\left(E_{e_{k}\longleftrightarrow i_{k}}^{s}\right)=\begin{cases}
1, & \textrm{if }\mathbb{I}_{\mathrm{node}}\left(i_{k}\right)=1\\
0, & \textrm{otherwise}
\end{cases}
\end{equation}
}{\footnotesize\par}

In the example shown in Figure \ref{fig:Graph-example}, the ego and
the first three actors are part of the graph with bidirectional spatial
edges (i.e., $d_{ei}$ is set as the edge feature) between the ego
and the actors. Furthermore, two of the detected actors $i_{k}$ and
$j_{k}$ share bidirectional edges between themselves if the euclidean
distance $d_{ij}$ between them is less than or equal to a predefined
threshold $d_{\mathrm{min}}=25\,\mathrm{m}$. {\footnotesize{}
\begin{equation}
\mathbb{I}_{\mathrm{aa}}\left(E_{i_{k}\longleftrightarrow j_{k}}^{s}\right)=\begin{cases}
1, & \textrm{if }\mathbb{I}_{\mathrm{node}}\left(i_{k}\right)=1\\
 & \textrm{and }\mathbb{I}_{\mathrm{node}}\left(j_{k}\right)=1\\
 & \textrm{and }d_{ij}\leq d_{\mathrm{min}}\\
0, & \textrm{otherwise}
\end{cases}\label{eq:edge-aa}
\end{equation}
}In the example shown in Figure \ref{fig:Graph-example}, while actors
1 and 2, and, 2 and 3 share an edge between themselves, actors 1 and
3 do not have an edge between them since $d_{13}>d_{\mathrm{min}}$.
Because of these spatial edges, $G_{k}$ is referred to as a spatial
graph.

Next, the temporal aspect is incorporated to generate the heterogeneous
graph $\mathcal{G}$. The temporal unidirectional edge is defined
as follows:

{\footnotesize{}
\begin{equation}
\mathbb{I}_{\mathrm{temp}}\left(E_{i_{k}\rightarrow i_{k+1}}^{t}\right)=\begin{cases}
1, & \textrm{if }\mathbb{I}_{\mathrm{node}}\left(i_{k}\right)=1\\
 & \textrm{and }\mathbb{I}_{\mathrm{node}}\left(i_{k+1}\right)=1\\
0, & \textrm{otherwise}
\end{cases},
\end{equation}
}i.e., if actor $i$ is within the sensing area during both the $k^{\mathrm{th}}$and
$\left(k+1\right)^{\mathrm{th}}$ time frames, they are connected
by a temporal unidirectional edge. Intuitively, the temporal edges
encapsulate the sequential aspect of the historical data $\mathcal{H}$.
The edge attribute is set to be the sampling time $t_{s}$. Needless
to say, the temporal edge for the ego always exists between the $k^{\mathrm{th}}$
and $\left(k+1\right)^{\mathrm{th}}$ time steps.

For example, as shown in the heterogeneous graph $\mathcal{G}$ of
Figure \ref{fig:Network-Architecture}, from $G_{1}$ to $G_{2}$,
there exist temporal edges for actors 1 and 2 since both these actors
were within the sensing area in both frames. However, actors 3 and
4 do not have the temporal edges since at $k=2$, the former is not
in the sensing area anymore while the latter materialized into the
sensing area for the first time. From $G_{2}$ to $G_{3}$, nothing
changed in the sensing area, and thus, all the three actors (1, 2
and 4) have temporal edge connections. Note that the two different
types of edges (spatial and temporal) in the formulation of $\mathcal{G}$
makes it a heterogeneous graph.

\subsection{Graph Encoder Network}

\begin{figure}[t]
\begin{centering}
\includegraphics[scale=0.07]{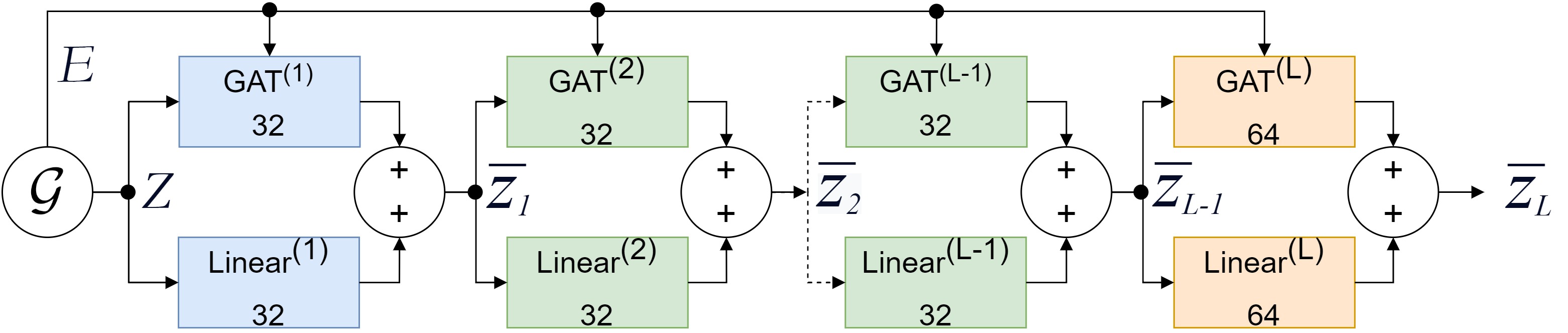}
\par\end{centering}
\begin{centering}
\vspace*{-3mm} 
\par\end{centering}
\caption{Graph encoder network with $L$ number of layers. GAT modules with
the same color share same weights and linear modules with the same
color share same weights. \label{fig:Graph-encoder-network}}
\end{figure}

The idea of the graph encoder network is to capture the context from
the history $\mathcal{H}$ presented as a spatial-temporal heterogeneous
graph $\mathcal{G}$. The graph encoder network consists of a series
of GAT convolutional networks in parallel with a series of linear
layers as shown in Figure \ref{fig:Graph-encoder-network}. The heterogeneous
graph $\mathcal{G}=\left(Z,E\right)$ is forwarded to get the final
encoded embedding $\bar{Z}_{L}$ of all the nodes in the graph, where
$Z$ is the feature array of all nodes in $\mathcal{G}$ and $E$
is the edge connection array between these nodes and the corresponding
edge attributes. If the total number of nodes in $\mathcal{G}$ is
$N_{\mathcal{G}}$, then $Z\in\mathbb{R}^{N_{\mathcal{G}}\times4}$
and thus, $\bar{Z}_{L}\in\mathbb{R}^{N_{\mathcal{G}}\times64}$ as
indicated by Figure \ref{fig:Graph-encoder-network} Formally, the
forward pass can be defined using the following equations:{\footnotesize{}
\begin{equation}
\begin{aligned}\bar{Z}_{1}= & \mathrm{GAT}^{(1)}\left(Z,E\right)+\mathrm{Linear}^{(1)}\left(Z\right)\\
\bar{Z}_{l}= & \mathrm{GAT}^{(l)}\left(\bar{Z}_{l-1},E\right)+\mathrm{Linear}^{(l)}\left(\bar{Z}_{l-1}\right),\\
 & \forall l=2,\ldots,L-1\\
\bar{Z}_{L}= & \mathrm{GAT}^{(L)}\left(\bar{Z}_{L-1},E\right)+\mathrm{Linear}^{(L)}\left(\bar{Z}_{L-1}\right)\\
= & \left[\begin{array}{cccc}
\bar{G}_{1} & \bar{G}_{2} & \ldots & \bar{G}_{K_{H}}\end{array}\right]
\end{aligned}
\end{equation}
}where, $\bar{G}_{k}$ contains the node embedding of the nodes of
the spatial graph $G_{k}$. In the above discussed example, $\bar{Z}_{L}\in\mathbb{R}^{12\times64}$
and $\bar{G}_{1},\bar{G}_{2},\bar{G}_{3}\in\mathbb{R}^{4\times64}$.
However, we are only interested in the embedding $\bar{G}_{K_{H}}\in\bar{Z}_{L}$
-- this fixed-size representation contains meaningful features about
the entire input sequence due to the unidirectional temporal edges.

\subsection{Sequential GRU Decoder Network}

The task is to predict the trajectory of actor $i_{K_{H}}\in\mathcal{N}_{K_{H}}$.
Without any loss of generality, for simplicity, we will refer to actor
$i_{K_{H}}$as $i$. The task of the decoder is to generate a decoded
state based on the current hidden state and the previously generated
decoded state. Needless to say, at the beginning, the sequential GRU
decoder network takes as inputs the initial hidden states $h_{K_{H}}^{i}$
(initialized as zeros) and {\footnotesize{}
\begin{equation}
\bar{z}_{K_{H}}^{i}=f\left(\bar{G}_{K_{H}},i\right)\in\bar{G}_{K_{H}}
\end{equation}
}where, $f\left(\bar{G}_{K_{H}},i\right)$ is a function that extracts
the corresponding embedding of actor $i$ from $\bar{G}_{K_{H}}$.
We also apply a residual connection similar to \cite{he2016deep}
to the output $g$ of the GRU as it allows to leverage previously
learned representations. Thus, the forward pass in the network can
be defined as follows:{\footnotesize{}
\begin{equation}
\begin{aligned}\left(g_{K_{H}+1}^{i},h_{K_{H}+1}^{i}\right)= & \mathrm{GRU}\left(\bar{z}_{K_{H}}^{i},h_{K_{H}}^{i}\right)\\
\left(g_{K_{H}+2}^{i},h_{K_{H}+2}^{i}\right)= & \mathrm{GRU}\left(\bar{z}_{K_{H}}^{i}+g_{K_{H}+1}^{i},h_{K_{H+1}}^{i}\right)\\
\left(g_{k}^{i},h_{k}^{i}\right)= & \mathrm{GRU}\left(g_{k-1}^{i}+g_{k-2}^{i},h_{k-1}^{i}\right),\\
 & \textrm{for }k=K_{H}+3,\ldots,K_{H}+K_{F}
\end{aligned}
\end{equation}
}{\footnotesize\par}

\subsection{Output Network}

For the prediction horizon $k=K_{H}+1,\ldots,K_{H}+K_{F}$, the output
network consists of a Multilayer perceptron (MLP) that takes $g_{k}^{i}$
as input to predict the trajectory of actor $i$. The output of the
MLP $\triangle Y_{k}^{i}$ is the change in position in both the $x$
and $y$ coordinates with respect to the previous actual or the estimated
position. Thus, the final predicted position $\hat{Y}_{k}^{i}$ for
the prediction horizon is obtained as follows:{\footnotesize{}
\begin{equation}
\begin{aligned}\hat{Y}_{K_{H}+1}^{i}= & \mathrm{MLP}\left(C\left(g_{K_{H}+1}^{i},Y_{K_{H}}^{i}\right)\right)+Y_{K_{H}}^{i}\\
\hat{Y}_{k}^{i}= & \mathrm{MLP}\left(C\left(g_{k}^{i},\hat{Y}_{k-1}^{i}\right)\right)+\hat{Y}_{k-1}^{i},\\
 & \textrm{for }k=K_{H}+2,\ldots,K_{H}+K_{F}
\end{aligned}
\label{eq:output_equation}
\end{equation}
}where, $Y_{K_{H}}^{i}$ is the current position of actor $i$ and
$C$ is a concatenation function. Note that AiGem does not simultaneously
predict trajectories of the $N_{K_{H}}$ actors, rather predicts trajectories
of each actor at a time. 

\section{Experiment Setup and Results}

\subsection{Datasets}

NGSIM datasets \cite{colyar2006next,colyar2007next}, collected by
the Federal Highway Administration of the U.S. Department of Transportation,
are used to evaluate the performance of our proposed model. In this
work, we use the datasets for highways US-101 \cite{colyar2007next}
and I-80 \cite{colyar2006next}. The NGSIM datasets consist of 45
minutes of vehicle trajectories transcribed from videos -- these
videos are obtained through synchronized cameras mounted on top of
adjacent buildings of the highway of interest. Many deep-learning
techniques \cite{deo2018convolutional,li2019grip++,lin2021vehicle,katariya2022deeptrack,sheng2022graph,alinezhad2023pishgu}
use these two datasets for performance assessment. 

\subsection{Data Processing}

The data in NGSIM were recorded at 10 frames per second, i.e., the
sampling time is 0.1 second. However, to make a fair comparison with
other techniques \cite{deo2018convolutional,li2019grip++,lin2021vehicle,katariya2022deeptrack,sheng2022graph}
which downsampled the data to 0.2 second, we do the same as well.
The trajectories are then segmented into 8 seconds blocks so that
the first 3 seconds can be used as historical observation, and the
remaining 5 seconds can be used as prediction ground truth. Similar
to \cite{lin2021vehicle,katariya2022deeptrack}, we split our dataset
into 70\% training data, 10\% validation data, and 20\% test data.
Note that in the NGSIM datasets, the heading values $\theta$ of the
vehicles are not provided. However, given the history of a vehicle's
trajectory, we can easily calculate $\theta$ for this past trajectory
using basic trigonometry. 

MinMax normalization is applied to all the inputs. The range for scaling
the positions and the heading is $\left(-1,1\right)$ and the range
for scaling the velocity and the distance between the vehicles (spatial
edge) is $\left(0,1\right)$. However, scaling is not applied on the
temporal edges since the magnitude of the sampling time (0.2 second)
is already within the range $\left(0,1\right)$.

\subsection{Baselines}

Recently, several deep-learning techniques have applied to conduct
the task of trajectory prediction on the NGSIM data. We use the following
models for comparing the performance of our proposed AiGem:

\noindent \textbf{(1) CS-LSTM }\cite{deo2018convolutional}\textbf{:
}It utilizes an LSTM encoder-decoder model with a social pooling layer
for feature extraction.

\noindent \textbf{(2) GRIP++} \cite{li2019grip++}\textbf{: }It utilizes
a LSTM encoder-decoder model with fixed and dynamic graphs to capture
the environmental dynamics.

\noindent \textbf{(3) STA-LSTM} \cite{lin2021vehicle}\textbf{: }It
combines spatial-temporal attention with LSTM and increases the interpretability
of the predictions.

\noindent \textbf{(4) DeepTrack} \cite{katariya2022deeptrack}\textbf{:
}It provides a light-weight prediction model by introducing temporal
and depthwise convolutions for capturing vehicle dynamics.

\noindent \textbf{(5) GSTCN} \cite{sheng2022graph}\textbf{: }It utilizes
a graph-based spatial-temporal convolutional network to first learn
the spatial features and then extract the temporal features. Finally,
GRU is used for prediction using the extracted features.

\subsection{Evaluation Metrics}

Our proposed model, AiGem, is compared against existing deep-learning
techniques using different metrics. The followings are commonly used
as a measure of prediction accuracy and performance of the system:\textbf{ }

\noindent \textbf{Average displacement error (ADE):} It is the average
euclidean distance between the predicted positions $\hat{Y}_{k}$
and the ground truth $Y_{k}$ for $k=K_{H}+1,\ldots,K_{H}+K_{F}$
and for $N_{K_{H}}$ actors:{\footnotesize{}
\begin{equation}
\mathrm{ADE}=\frac{\sum_{n=1}^{N_{K_{H}}}\sum_{k=K_{H}+1}^{K_{H}+K_{F}}\left\Vert \hat{Y}_{k}^{i}-Y_{k}^{i}\right\Vert _{2}}{K_{F}N_{K_{H}}}
\end{equation}
}{\footnotesize\par}

\noindent \textbf{Final displacement error (FDE):} It is the average
euclidean distance between $\hat{Y}_{k}$ and $Y_{k}$ for the last
predicted step $k=K_{H}+K_{F}$ and for $N_{K_{H}}$ actors:{\footnotesize{}
\begin{equation}
\mathrm{FDE}=\frac{\sum_{n=1}^{N_{K_{H}}}\left\Vert \hat{Y}_{K_{H}+K_{F}}^{i}-Y_{K_{H}+K_{F}}^{i}\right\Vert _{2}}{N_{K_{H}}}
\end{equation}
}{\footnotesize\par}

\noindent \textbf{Root mean square error (RMSE):} It is the square
root of the mean sequared error between $\hat{Y}_{k}$ and $Y_{k}$
at $k^{\mathrm{th}}$ step for $N_{K_{H}}$ actors:{\footnotesize{}
\begin{equation}
\mathrm{RMSE}=\sqrt{\frac{1}{N_{K_{H}}}\stackrel[n=1]{N_{K_{H}}}{\sum}\left(\hat{Y}_{k}^{i}-Y_{k}^{i}\right)^{2}}
\end{equation}
}We also compare the model size (i.e., number of parameters in the
model) of AiGem with the baselines.

\subsection{Ablation Study}

We conduct two ablation studies, particularly, (1) the effect of $d_{\mathrm{min}}$
as described in equation (\ref{eq:edge-aa}), and, (2) the effect
of concatenation at the MLP input defined in equation (\ref{eq:output_equation})
on the performance of the AiGem.

\begin{figure}[t]
\begin{centering}
\includegraphics[scale=0.275]{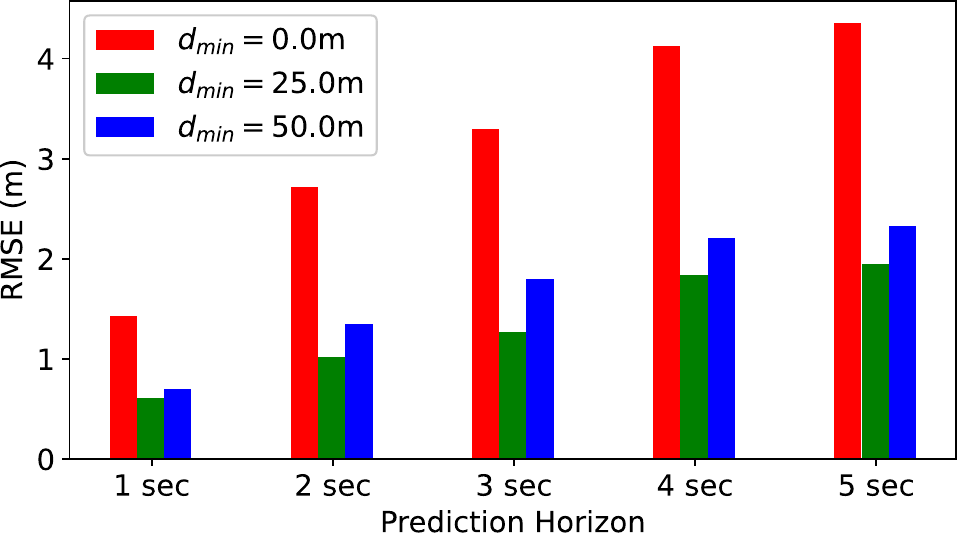}
\par\end{centering}
\begin{centering}
\vspace*{-3mm} 
\par\end{centering}
\caption{Comparison of performance of AiGem for different values of $d_{\mathrm{min}}$
\label{fig:Comparison-dmin}}
\end{figure}

\textbf{(1)} For the formulation of the graph in section \ref{subsec:Graph-Formation},
equation (\ref{eq:edge-aa}) defines that there exists a bidirectional
edge between two actors if the distance between them is less than
a predefined threshold $d_{\mathrm{min}}$. We set the values of $d_{\mathrm{min}}$
to different values and compare the performance of AiGem for these
different values. Figure \ref{fig:Comparison-dmin} shows how our
proposed model performs for different values of $d_{\mathrm{min}}$.
It is clear that when $d_{\mathrm{min}}=0\textrm{ m}$(i.e., none
of the detected actors are connected with each other using an edge
in the spatial graph), the error is maximum for all the prediction
horizons. Therefore, connecting actors using edges in the spatial
graph indeed help minimizing the loss. Next, it can be observed that
the error also increases when $d_{\mathrm{min}}$ is increased from
$\textrm{25 m}$ to $\textrm{50 m}$, i.e., the number of bidirectional
edges between actors increases due to increase in $d_{\mathrm{min}}$.
This phenomenon makes intuitive sense since in real-life scenarios,
a human driver is more likely to make decisions based on nearer actors
than actors that are further. Since $d_{\mathrm{min}}=\textrm{25 m}$
results in the minimum RMSE value, we use that in the formulation
of the spatial graphs.

\begin{figure}[t]
\begin{centering}
\includegraphics[scale=0.25]{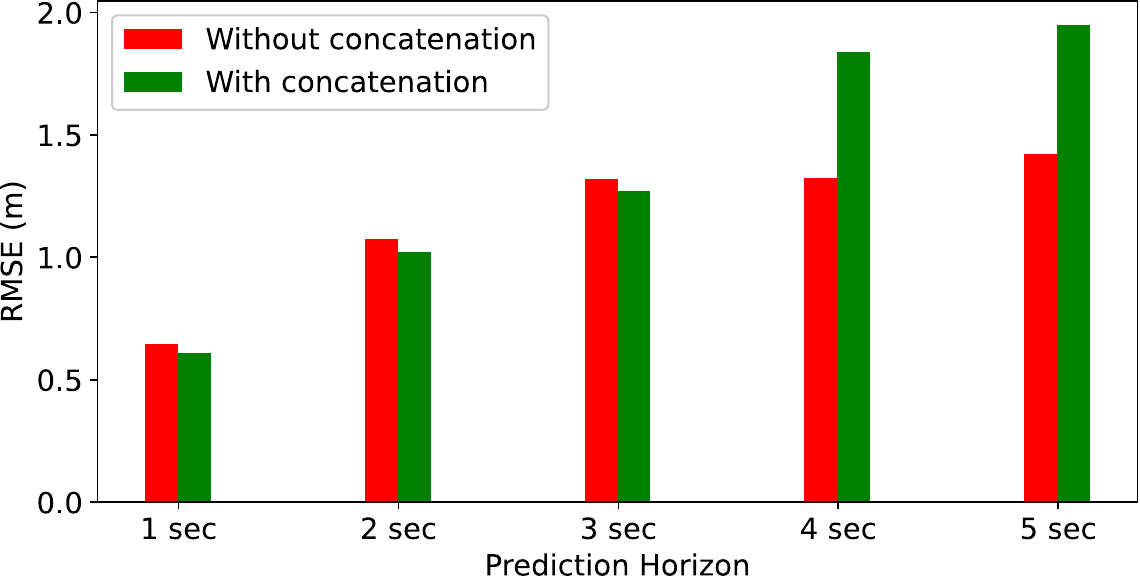}
\par\end{centering}
\begin{centering}
\vspace*{-3mm} 
\par\end{centering}
{\small{}\caption{Comparison of performance of AiGem with and without concatenation
of the output\label{fig:Comparison-concat} }
}{\small\par}

\end{figure}

\textbf{(2)} In the proposed architecture, as defined in equation
(\ref{eq:output_equation}), we concatenate the output with the decoded
embedding at the input of the MLP of the output network. We want to
see the performance without concatenation, i.e., modify equation (\ref{eq:output_equation})
to:{\footnotesize{}
\begin{equation}
\begin{aligned}\hat{Y}_{K_{H}+1}^{i}= & \mathrm{MLP}\left(g_{K_{H}+1}^{i}\right)+Y_{K_{H}}^{i}\\
\hat{Y}_{k}^{i}= & \mathrm{MLP}\left(g_{k}^{i}\right)+\hat{Y}_{k-1}^{i},\\
 & \textrm{for }k=K_{H}+2,\ldots,K_{H}+K_{F}
\end{aligned}
\label{eq:output_equation-1}
\end{equation}
}{\small{}Figure \ref{fig:Comparison-concat} shows the differences
between the performances with and without concatenation for different
prediction horizons. It is clear that for lower prediction horizons
(1, 2 and 3 seconds), concatenation has clear advantage. However,
for longer predictions (4 and 5 seconds), concatenation degrades the
performance. Since concatenation has advantage over three prediction
horizons out of five, we use the concatenation approach to report
in the remaining article.}
{\small\par}

\subsection{AiGem Training}

\begin{figure*}[t]
\begin{centering}
\includegraphics[scale=0.25]{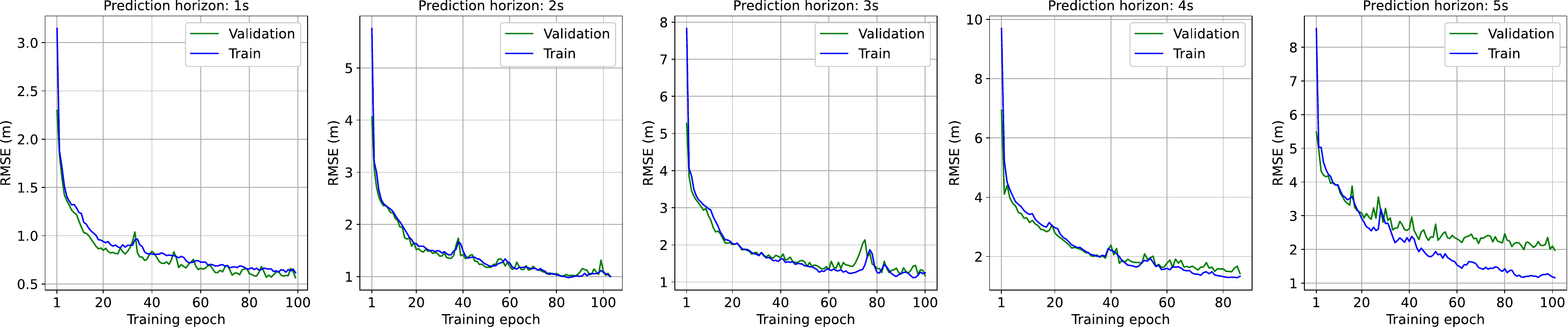}
\par\end{centering}
\begin{centering}
\vspace*{-3mm} 
\par\end{centering}
\caption{Training of separate AiGem models for different prediction horizons
showing RMSE evolution for training and validation datasets \label{fig:Training-of-AiGem}}
\end{figure*}

In our work, we used the Adam optimizer to optimize AiGem with a learning
rate of 0.001. To mitigate overfitting of the network and improve
generalization performance of our proposed network, dropout regularization
technique is applied during training, i.e., a dropout rate of 5\%
was applied. We train a separate model for each target horizon. Figure
\ref{fig:Training-of-AiGem} shows the training evolution of the AiGem
models for different prediction horizons. Note that it took approximately
100 training epochs to get the best model for each prediction horizon
-- the best model refers to the model with the lowest validation
loss. As the prediction horizon increases, overfitting becomes evident
in the training of AiGem.

\subsection{Results}

\begin{figure}[t]
\begin{centering}
\includegraphics[scale=0.275]{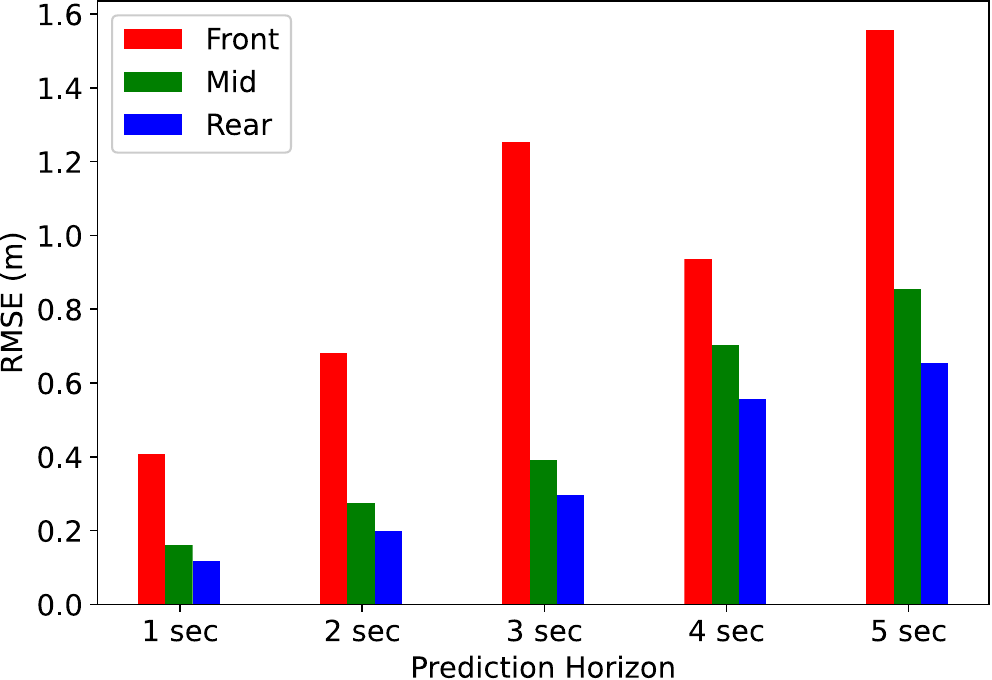}
\par\end{centering}
\begin{centering}
\vspace*{-3mm} 
\par\end{centering}
\caption{Performance assessment of AiGem for predicting trajectories of actors
based on the positions around the ego \label{fig:Performance-Actor-Position}}
\end{figure}

Table \ref{tab:Comparison-Table} shows the performance comparison
between AiGem and other baselines. In RMSE comparison, AiGem achieves
comparable results in short predictions, i.e., the first two seconds.
However, in the longer prediction horizons, it is at the forefront
of all the models. For 3 seconds and 4 seconds prediction, it achieves
a lead of 4.5\% and 8.5\% from the second best, respectively. The
lead becomes significant for 5 seconds -- 33.7\% improvement from
the second best. This shows that AiGem is more proficient in extracting
useful features for longer predictions.

As shown in Table \ref{tab:Comparison-Table}, compared to CS-LSTM,
GRIP++, STA-LSTM, DeepTrack and GSTCN, AiGem reduces ADE by 38.0\%,
11.8\%, 24.8\%, 29.4\% and 6.6\%, respectively. For the metric FDE,
AiGem has the best score among all the models. It outperforms the
baselines by 23.1\%, 18.7\% and 20.9\% compared to CS-LSTM, STA-LSTM
and DeepTrack, respectively.

The AiGem is also the second lightest model in terms of model size
as shown in Table \ref{tab:Comparison-Table}. Compared to heaviest
model GRIP++, it has 85.5\% less number of parameters. The lightest
model is the GSTCN with 33.2\% less parameters than ours. The smaller
size of GSTCN is likely due to restricting the graph to two laterally
adjacent lanes and $\pm100$ meters of the roadways \cite{alinezhad2023pishgu}
in contrary to AiGem. As explained earlier in section \ref{sec:Problem-Formulation},
the shifting of the coordinate frame with respect to the current position
of the ego allows our model to adjust easily to any section of the
road.

In summary, our proposed model AiGem is lightweight yet excels in
forecasting trajectories over longer prediction horizons, outperforming
baseline models in both longer horizon predictions, the ADE score
and the FDE score. 
\begin{table}[t]
\caption{Prediction metrics and model size comparison of AiGem with the baselines.
The best results are in bold and the second best are underlined. \label{tab:Comparison-Table}}

\begin{centering}
\begin{tabular}{>{\raggedright}m{0.4in}>{\raggedright}m{0.2in}>{\raggedright}m{0.2in}ccccc>{\raggedright}m{0.275in}}
\toprule 
\multirow{2}{0.4in}{\textbf{\scriptsize{}Model}} & \multirow{2}{0.2in}{\textbf{\scriptsize{}ADE (m)}} & \multirow{2}{0.2in}{\textbf{\scriptsize{}FDE (m)}} & \multicolumn{5}{c}{\textbf{\scriptsize{}RMSE (m)}} & \multirow{2}{0.275in}{\textbf{\scriptsize{}Params (K)}}\tabularnewline
\cmidrule{4-8} \cmidrule{5-8} \cmidrule{6-8} \cmidrule{7-8} \cmidrule{8-8} 
 &  &  & \textbf{\scriptsize{}1s} & \textbf{\scriptsize{}2s} & \textbf{\scriptsize{}3s} & \textbf{\scriptsize{}4s} & \textbf{\scriptsize{}5s} & \tabularnewline
\midrule 
\textbf{\scriptsize{}CS-LSTM} & {\scriptsize{}2.29} & {\scriptsize{}3.34} & {\scriptsize{}0.61} & {\scriptsize{}1.27} & {\scriptsize{}2.09} & {\scriptsize{}3.10} & {\scriptsize{}4.37} & {\scriptsize{}192}\tabularnewline
\midrule 
\begin{raggedright}
{\scriptsize{}\vspace{0.1cm}
}{\scriptsize\par}
\par\end{raggedright}
\textbf{\scriptsize{}GRIP++}{\scriptsize{}\vspace{0.1cm}
}\textbf{\scriptsize{} } & {\scriptsize{}1.61} & {\scriptsize{}--} & {\scriptsize{}\uline{0.38}} & {\scriptsize{}\uline{0.89}} & {\scriptsize{}1.45} & {\scriptsize{}2.14} & {\scriptsize{}\uline{2.94}} & {\scriptsize{}496{*}}\tabularnewline
\midrule 
\textbf{\scriptsize{}STA-LSTM} & {\scriptsize{}1.89} & {\scriptsize{}\uline{3.16}} & \textbf{\scriptsize{}0.37} & {\scriptsize{}0.98} & {\scriptsize{}1.71} & {\scriptsize{}2.63} & {\scriptsize{}3.78} & {\scriptsize{}125}\tabularnewline
\midrule 
\textbf{\scriptsize{}Deep Track} & {\scriptsize{}2.01} & {\scriptsize{}3.25} & {\scriptsize{}0.47} & {\scriptsize{}1.08} & {\scriptsize{}1.83} & {\scriptsize{}2.75} & {\scriptsize{}3.89} & {\scriptsize{}109}\tabularnewline
\midrule 
\begin{raggedright}
{\scriptsize{}\vspace{0.1cm}
}{\scriptsize\par}
\par\end{raggedright}
\textbf{\scriptsize{}GSTCN}{\scriptsize{}\vspace{0.1cm}
} & {\scriptsize{}\uline{1.52}} & {\scriptsize{}--} & {\scriptsize{}0.44} & \textbf{\scriptsize{}0.83} & {\scriptsize{}\uline{1.33}} & {\scriptsize{}\uline{2.01}} & {\scriptsize{}2.98} & \textbf{\scriptsize{}49.8}\tabularnewline
\midrule 
\textbf{\scriptsize{}AiGem (Ours)} & \textbf{\scriptsize{}1.42} & \textbf{\scriptsize{}2.57} & {\scriptsize{}0.61} & {\scriptsize{}1.02} & \textbf{\scriptsize{}1.27} & \textbf{\scriptsize{}1.84} & \textbf{\scriptsize{}1.95} & {\scriptsize{}\uline{74.5}}\tabularnewline
\bottomrule
\end{tabular}
\par\end{centering}
\begin{raggedright}
{\scriptsize{}\vspace{0.1cm}
{*}\cite{li2019grip++} does not report the number of parameters but
we extracted this number from}{\scriptsize\par}
\par\end{raggedright}
\raggedright{}{\scriptsize{}their shared code \cite{grip++code}}{\scriptsize\par}
\end{table}

Furthermore, we analyze the performance of AiGem on predicting trajectories
of actors based on their positions around the ego. To do that, we
broadly define three positions:
\begin{enumerate}
\item \textbf{Front position}: If an actor is more than 15 m ahead longitudinally
from the ego, it is considered in the front position.
\item \textbf{Rear position}: If an actor is more than 15 m behind longitudinally
from the ego, it is considered in the rear position.
\item \textbf{Mid position}: If an actor is between $\pm$15 m longitudinally
of the ego, it is considered in the mid position.
\end{enumerate}
Figure \ref{fig:Performance-Actor-Position} shows the performance
of AiGem in predicting trajectories of actors based on the positions
described above. It can be clearly observed that our proposed model
is able to predict trajectories of actors more accurately that are
in rear position with respect to the ego, in contrast to the mid and
front positions, for all prediction horizons. The model achieves the
least accuracy when the actors are in the front position. Thus, it
can be concluded, when AiGem predicts the trajectory of an actor $i$
with respect to the ego, it considers front actors relative to $i$
to have more impact on its future decision-making. On the other hand,
rear actors relative to $i$ has lesser impact on its future decision-making.
It is important to note that the graph formulation is constrained
by the sensing area of the ego -- this implies that the actors positioned
in the rear with respect to the ego observes actors in front of them
while the actors positioned in the front with respect to the ego observes
actors behind them.

\section{Conclusion }

In this article, we propose a deep learning model called the AiGem
that constructs a heterogeneous graph from the historical data using
spatial and temporal edges to capture interactions between the agents.
A graph encoder network generates embedding for the target actors
in the current timestamp which is fed through a sequential GRU decoder
network. The decoded states from the decoder network are then utilized
to predict future trajectories using an MLP in the output network.
NGSIM datasets have been used for performance assessment. The results
show that AiGem achieves comparable accuracy to state-of-the-art prediction
algorithms for shorter prediction horizons. For longer prediction
horizons, particularly 3, 4 and 5 seconds, it outperforms all the
baselines used. The size of the model is better than most of them
and comparable to the lightest.

\bibliographystyle{IEEEtran.bst}
\bibliography{ReferencesTrajectoryPrediction}

\end{document}